\documentclass[manuscript]{acmart}

\usepackage{booktabs}
\usepackage{multirow}
\usepackage{graphicx}
\usepackage{tikz}
\usetikzlibrary{arrows.meta,positioning}

\title{Finding Is Not Returning, Returning Is Not Reconstructing: \\A Multi-Dimensional Evaluation of Agent Social Re-entry}

\author{Sheng Wan}
\authornote{Both authors contributed equally to this research.}
\email{swanae@connect.ust.hk}
\affiliation{%
  \institution{The Hong Kong University of Science and Technology}
  \city{Hong Kong}
  \country{China}
}
\affiliation{%
  \institution{Southern University of Science and Technology}
  \city{Shenzhen}
  \country{China}
}

\author{Cong Ma}
\authornotemark[1]
\affiliation{%
  \institution{Southern University of Science and Technology}
  \city{Shenzhen}
  \country{China}
}
\affiliation{%
  \institution{City University of Hong Kong}
  \city{Hong Kong}
  \country{China}
}

\author{Sicheng Dong}
\affiliation{%
  \institution{Shenzhen MSU-BIT University}
  \city{Shenzhen}
  \country{China}
}

\author{Qian'ang Mao}
\affiliation{%
  \institution{Nanjing University}
  \city{Nanjing}
  \country{China}
}

\author{Chenghan Zhang}
\affiliation{%
  \institution{University of Technology Sydney}
  \city{Sydney}
  \country{Australia}
}

\author{Daning Hu}
\affiliation{%
  \institution{Southern University of Science and Technology}
  \city{Shenzhen}
  \country{China}
}

\author{Qiang Yang}
\authornote{Corresponding author.}
\email{profqiang.yang@polyu.edu.hk}
\affiliation{%
  \institution{The Hong Kong Polytechnic University}
  \city{Hong Kong}
  \country{China}
}

\acmJournal{TIST}

\begin{document}
\raggedbottom

\begin{abstract}
As Large Language Model (LLM) agents increasingly participate in persistent social environments, understanding their long-term social interaction capabilities becomes increasingly important. While existing research has advanced long-term memory and conversational context retrieval, less attention has been paid to whether agents can autonomously re-engage with previously inactive conversations. In particular, remembering a past interaction does not necessarily mean that an agent can identify when to revisit it, decide to return, or reconstruct the context needed for a coherent continuation. To investigate this gap, we introduce agent social re-entry as a multi-dimensional evaluation problem, distinguishing three complementary capabilities: Finding, Returning, and Reconstructing. We develop a shared-snapshot evaluation framework that measures these dimensions independently, allowing us to examine how conversational resurfacing and persistent memory relate to different aspects of social re-entry. Evaluations across six LLM configurations show that resurfacing substantially improves target conversation identification, while persistent memory is associated with higher rates of autonomous target-directed replying. However, neither improvement is accompanied by consistent gains in contextual reconstruction. These results highlight the distinction between accessing previous conversations, choosing to re-engage, and maintaining conversational continuity. Our framework provides a way to evaluate these capabilities separately rather than relying on memory retrieval or conversational response quality alone.
\end{abstract}


\maketitle

\section{Introduction}
\label{sec:introduction}

In online communities, participants often revisit earlier conversations after hours or days of inactivity. This behavior, known as conversational re-entry, has been studied in human online interactions~\cite{wang2021reentry}. With LLM agents increasingly operating in persistent social environments~\cite{park2023generative,xi2023agents,wang2024survey}, conversational re-entry also becomes relevant to autonomous agents. While prior research has investigated whether human participants will return to earlier discussions, an important question arises for LLM agents: \emph{Can they autonomously re-enter previous social conversations while maintaining contextual continuity?}

Recent advances in long-context modeling, retrieval-augmented generation, and agent memory have improved the ability of language models to retain and access previous information~\cite{lewis2020rag,liu2023lost}. However, these advances primarily address the availability and use of historical information, whereas social re-entry also requires an agent to decide when and where to resume a previous interaction and to maintain continuity with that interaction. Consequently, evaluating retrieval accuracy or conversational memory alone is insufficient to determine whether an agent can successfully re-engage with previous social interactions.

We formulate agent social re-entry as a multi-dimensional evaluation problem involving three conceptually connected stages: \emph{Finding}, \emph{Returning}, and \emph{Reconstructing}. Finding concerns identifying a relevant prior conversation. Returning concerns autonomously re-engaging with that conversation rather than merely recognizing it. Reconstructing concerns recovering the relevant conversational context to produce a coherent continuation. These stages form a natural conceptual progression, but success at one stage does not guarantee success at the next. An agent may find a previous conversation without returning to it, or return without successfully reconstructing its context. Related research on conversational grounding, interruption and resumption, and prospective memory provides useful perspectives on these challenges~\cite{clark1991grounding,mcfarlane2002interruption,mark2008cost,einstein1990prospective}. Thus, \emph{Finding does not guarantee Returning, and Returning does not guarantee Reconstructing.}

To systematically evaluate these stages, we propose a controlled evaluation framework for agent social re-entry. Rather than relying solely on end-to-end observations, the framework evaluates each stage through a separate experimental branch initialized from a common world-state snapshot. Finding is measured by selecting a target conversation among ten candidate threads, Returning by whether an agent produces an unprompted, target-directed reply, and Reconstructing by the contextual continuity of a prompted continuation. This design allows the stages to be evaluated separately without actions in one branch influencing measurements in another. We further investigate persistent memory and prioritized resurfacing through a $2\times2$ experimental design, supplemented by distractor-memory and oracle controls. The framework is evaluated across six service/model configurations under a common experimental protocol.

Our results show distinct patterns across the three stages. Prioritized resurfacing substantially improves Finding, while the persistent-memory configuration is associated with higher Returning rates. However, corresponding improvements in Reconstructing are not observed. These patterns are consistent across the six evaluated configurations, although the reconstruction differences vary in magnitude and statistical significance. Together, these findings highlight the importance of evaluating Finding, Returning, and Reconstructing separately rather than treating them as a single measure of social re-entry.

This work makes three main contributions. First, we formulate \textbf{agent social re-entry} as a multi-dimensional evaluation problem, distinguishing Finding, Returning, and Reconstructing as conceptually connected but separately measurable capabilities. Second, we propose a shared-snapshot evaluation framework that independently measures these dimensions and supports controlled interventions on persistent memory and conversational resurfacing. Third, through experiments across six service/model configurations, we find that \textbf{successful Finding does not necessarily imply Returning, and increased Returning does not guarantee stronger Reconstructing}. These findings highlight the distinction between the three dimensions and the importance of evaluating them separately.

\section{Related Work}
\label{sec:related}

\paragraph{Conversational re-entry and social continuity.}
Conversational re-entry has been studied in human online communities, where the central task is to predict whether a participant will return to a conversation in which they previously participated. Wang et al.~\cite{wang2021reentry} investigate this problem by jointly modeling conversation context and user interaction history. Related research on conversational grounding emphasizes the establishment and maintenance of shared understanding~\cite{clark1991grounding}, while studies of interruptions and prospective memory examine the challenges of resuming previous activities and acting on delayed intentions~\cite{mcfarlane2002interruption,mark2008cost,einstein1990prospective}.

Recent studies have further examined conversational persistence and social interaction in agent-oriented environments. Eziz~\cite{eziz2026fast} analyzes reply dynamics and conversational persistence in Moltbook, highlighting the limited extent of sustained interactions. Zerhoudi et al.~\cite{zerhoudi2026form} investigate agent social behavior in Moltbook, reporting shallow conversational structures and infrequent returns by original post authors. These studies characterize observed interaction patterns in agent social networks, while our work evaluates agent social re-entry through three separately measurable capabilities: Finding, Returning, and Reconstructing.

\paragraph{Memory and context in LLM agents.}
Transformer-based language models generate responses conditioned on their input context~\cite{vaswani2017attention}, while retrieval-augmented generation incorporates externally retrieved information~\cite{lewis2020rag}. Persistent social-agent architectures further support the storage and retrieval of observations, reflections, and previous interactions~\cite{park2023generative}. However, access to additional context does not necessarily ensure its effective use, as demonstrated by the sensitivity of long-context models to the placement of relevant information~\cite{liu2023lost}. Recent benchmarks such as LoCoMo~\cite{maharana2024evaluating} evaluate long-term conversational memory through question answering, event summarization, and dialogue generation. These approaches provide important foundations for accessing and using conversational history. Our work examines a complementary question: whether mechanisms that make earlier conversations accessible also affect an agent's autonomous decision to return, and whether returning is accompanied by successful contextual reconstruction.

\paragraph{Evaluation of autonomous LLM agents.}
Research on LLM agents has investigated reasoning, planning, and action, including the use of external tools~\cite{yao2023react}. Surveys identify memory, planning, and action as central components of autonomous-agent systems~\cite{xi2023agents,wang2024survey}, while general evaluation frameworks emphasize the importance of assessing multiple aspects of model behavior~\cite{liang2023helm}. More specifically, SocialBench~\cite{chen2024socialbench} evaluates the social interaction capabilities of role-playing agents at individual and group levels. ChronosBench~\cite{shi2026longterm} investigates proactive long-term task-oriented interaction, including intent-conditioned monitoring and event-triggered follow-up. These studies extend agent evaluation beyond conventional task execution, but address different settings from resuming previously inactive social conversations. Our work focuses on agent social re-entry as a multi-stage process, distinguishing the ability to identify a relevant prior conversation, autonomously return to it, and reconstruct its context. By evaluating these stages through separate branches initialized from a common world-state snapshot, we examine their responses to memory and resurfacing interventions without assuming that success in one stage establishes successful end-to-end re-entry.

\begin{figure}[t]
\centering
\includegraphics[width=\columnwidth]{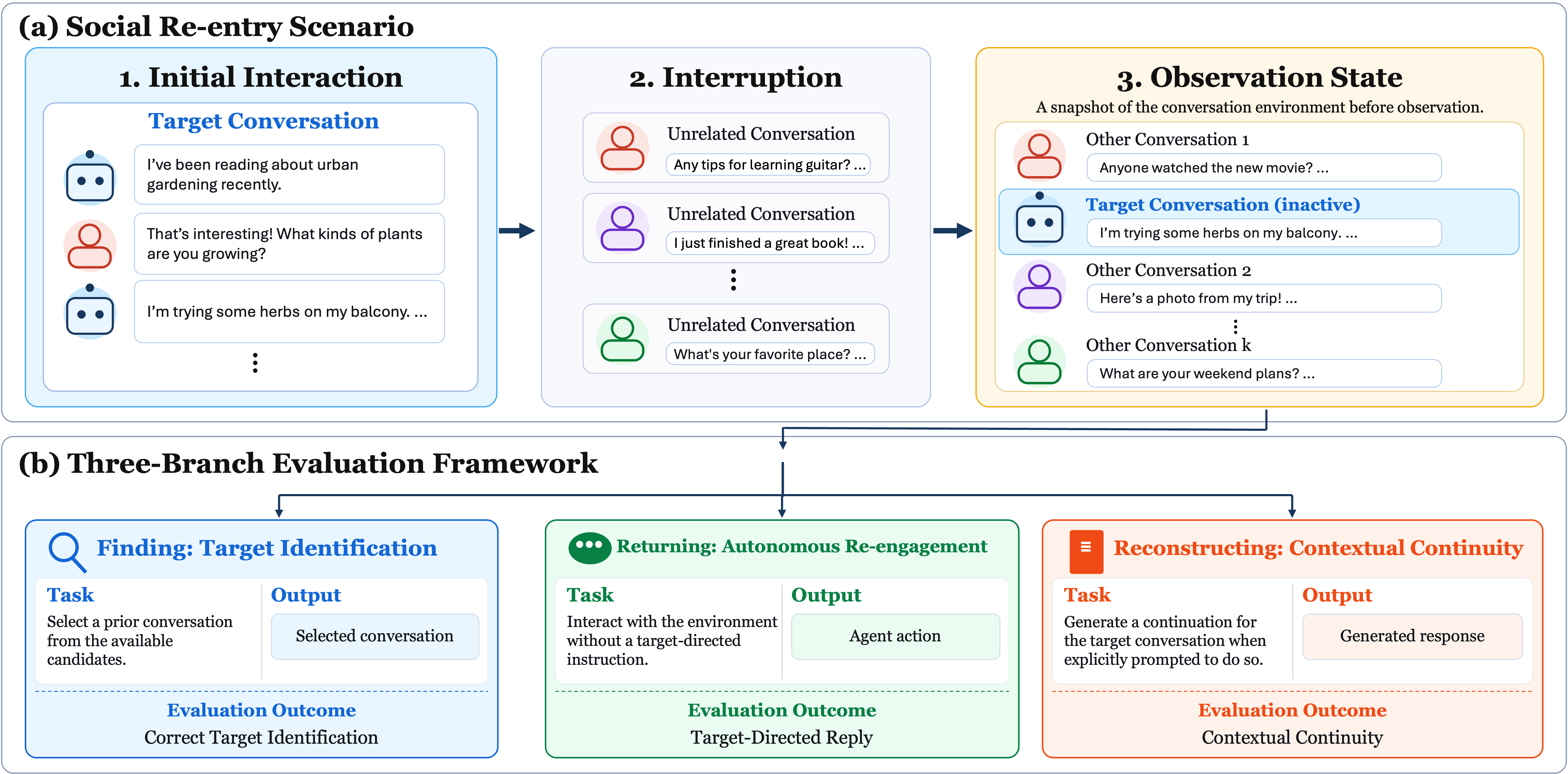}
\Description{Schematic overview of the agent social re-entry evaluation framework, consisting of two panels. Panel (a) illustrates an initial interaction that establishes a target conversation, followed by unrelated social activity during an interruption period and a frozen observation state. Panel (b) shows three independent evaluation branches initialized from copies of the same observation state. The Finding branch assesses target conversation identification, the Returning branch assesses autonomous target-directed replying, and the Reconstructing branch assesses contextual continuity in a prompted continuation. The branches are evaluated independently rather than sequentially.}
\caption{Overview of the agent social re-entry evaluation framework. (a) An initial interaction establishes a target conversation, followed by unrelated conversational activity during an interruption period. The resulting environment state is frozen before observation. (b) Three independent evaluation branches are initialized from copies of the same observation state to assess Finding, Returning, and Reconstructing under their respective task settings. The branches are evaluated separately rather than as a single sequential process.}
\label{fig:overview}
\end{figure}

\section{Method}
\label{sec:method}

\subsection{Problem Formulation}
\label{subsec:problem_formulation}

We study \emph{agent social re-entry}, which refers to the process by which an LLM agent identifies and autonomously returns to a previously inactive social conversation. Successful contextual continuation further requires the agent to reconstruct the relevant conversational history and produce a response that maintains continuity with the earlier interaction. Unlike conventional conversational memory evaluations, which assess whether models can retrieve and use information from previous interactions through explicitly specified tasks~\cite{maharana2024evaluating}, agent social re-entry additionally involves the decision to resume a previous interaction and the ability to continue it coherently.

Consider a social environment in which an agent $A$ has previously participated in a conversation $T^*$ with another participant. Following a period of inactivity, the agent encounters a set of available conversations $\mathcal{T}=\{T_1,\ldots,T_N\}$, where $T^*\in\mathcal{T}$ denotes the target conversation and the remaining threads represent competing conversational opportunities. The agent may have access to information about its earlier interactions, depending on the available conversational context and memory configuration. The central question is whether the agent can recognize the relevant prior conversation, choose to return to it without an explicit target-directed instruction, and maintain contextual continuity when continuing the discussion.

We conceptualize this process through three stages: \emph{Finding}, \emph{Returning}, and \emph{Reconstructing}. Finding concerns identifying the target conversation among available threads. Returning concerns autonomously re-engaging with the target conversation rather than merely recognizing it. Reconstructing concerns recovering the relevant conversational context and producing a coherent continuation. These stages form a natural conceptual progression: an agent must identify a conversation before deliberately returning to it, while a successful return does not necessarily imply that its continuation preserves the earlier context.

Importantly, the three stages describe distinct observable capabilities rather than independent internal cognitive mechanisms. Their conceptual relationship does not imply that they must be measured through a single sequential trajectory. Instead, we develop an evaluation framework that examines each stage separately, allowing us to distinguish target identification, autonomous returning, and contextual reconstruction without conflating their outcomes.

\subsection{Social Re-entry Evaluation Framework}
\label{subsec:evaluation_framework}

We introduce a controlled evaluation framework that represents agent social re-entry as an interrupted interaction followed by an opportunity for renewed engagement. As illustrated in Figure~\ref{fig:overview}, the framework consists of two components: (i) a social re-entry scenario involving an initial interaction, an interruption period, and a shared observation state (Figure~\ref{fig:overview}(a)); and (ii) three independent evaluation branches initialized from that state to assess Finding, Returning, and Reconstructing (Figure~\ref{fig:overview}(b)).

\paragraph{Initial interaction and interruption.}
The environment first establishes an interaction between the agent and another participant in a target conversation $T^*$. This interaction provides the conversational history that the agent may later need to recognize and reconstruct. The original interaction is followed by unrelated social activity, during which the target conversation remains inactive. The interruption separates the initial interaction from the subsequent opportunity for re-entry and introduces competing conversational context.

\paragraph{Shared observation state.}
Following the interruption, the environment reaches an observation state $S$, which contains the target conversation and other available conversations, as illustrated in Figure~\ref{fig:overview}(a). The state is frozen immediately before observation and serves as the common starting point for the three evaluation branches. Importantly, the shared state does not imply identical inputs across branches; the information and conversational cues presented to the agent depend on the corresponding evaluation task and experimental conditions.

\paragraph{Independent evaluation branches.}
Rather than evaluating all three stages through a single sequence of agent actions, we initialize three independent branches from the same observation state:

\begin{itemize}
    \item \emph{Finding branch}: evaluates whether the agent can identify the target conversation when explicitly asked to select a relevant prior thread.
    \item \emph{Returning branch}: evaluates whether the agent spontaneously produces a target-directed reply while interacting with the social environment, without being explicitly instructed to return to the target conversation.
    \item \emph{Reconstructing branch}: evaluates the contextual continuity of a response when the agent is explicitly prompted to continue the target conversation.
\end{itemize}

Each branch operates on an independent copy of the observation state. Consequently, actions performed in one branch do not alter the environment or conversational history used by another branch. This separation prevents an agent's behavior in one evaluation task from influencing the measurements of the others.

The framework distinguishes a conceptual progression from an experimental sequence. Finding, Returning, and Reconstructing represent connected aspects of successful social re-entry, but their measured outcomes are obtained independently. Therefore, the framework evaluates stage-specific capabilities rather than directly estimating end-to-end social re-entry success.

\subsection{Three Dimensions of Social Re-entry}
\label{subsec:three_dimensions}

As illustrated in Figure~\ref{fig:overview}(b), we operationalize Finding, Returning, and Reconstructing through three independent evaluation tasks, each with a distinct output and evaluation outcome. Let $T^*$ denote the target conversation, $\mathcal{T}$ the set of candidate conversations, and $S$ the shared observation state. The following definitions specify what each dimension measures.

\paragraph{Finding: target identification.}
Finding evaluates whether an agent can recognize the target conversation among available candidates. Given the candidate set $\mathcal{T}$, the agent is explicitly asked to select a conversation or abstain. Let $\hat{T}$ denote the selected conversation. The Finding outcome is defined as

\begin{equation}
F = \mathbb{I}[\hat{T}=T^*],
\label{eq:finding}
\end{equation}

where $\mathbb{I}[\cdot]$ is the indicator function. A successful Finding outcome indicates that the agent selects the target conversation. Because this task explicitly requests a selection, it measures target identification under instruction rather than the agent's spontaneous decision to revisit the conversation.

\paragraph{Returning: autonomous re-engagement.}
Returning evaluates whether the agent independently chooses to resume the target conversation during an observation opportunity. Unlike Finding, the agent is not explicitly instructed to identify or reply to $T^*$. Instead, it acts within the social environment, where multiple conversational opportunities are available.

Let $a$ denote the agent's observed action, and let $\operatorname{target}(a)$ and $\operatorname{text}(a)$ denote the destination thread and textual content of a reply action, respectively. We define the Returning outcome as

\begin{equation}
R = \mathbb{I}\left[
\operatorname{type}(a)=\mathrm{reply}
\;\land\;
\operatorname{target}(a)=T^*
\;\land\;
|\operatorname{text}(a)|>0
\right].
\label{eq:returning}
\end{equation}

A successful Returning outcome indicates that the agent produces a non-empty, target-directed reply without an explicit instruction to resume that particular conversation. This measure captures observable autonomous re-engagement, but does not establish whether the response accurately reconstructs the earlier discussion.

\paragraph{Reconstructing: contextual continuity.}
Reconstructing evaluates whether an agent can produce a continuation that reflects the relevant history of the target conversation. In this task, the agent is explicitly prompted to continue $T^*$, allowing contextual reconstruction to be assessed independently of its spontaneous decision to return.

Let $H^*$ denote the relevant history of the target conversation and $y$ the generated continuation. We define the Reconstructing outcome as

\begin{equation}
Q = J(y,H^*),
\qquad Q\in\{0,1,2,3\},
\label{eq:reconstructing}
\end{equation}

where $J$ denotes an LLM-based evaluation procedure that assigns an ordinal contextual continuity score. Higher scores indicate stronger continuity with the previous interaction, ranging from no meaningful continuity to a specific and natural continuation of the discussion.

Unlike Finding and Returning, which are measured as binary outcomes, Reconstructing is evaluated through an ordinal quality score. The three measures therefore capture different properties of agent behavior and should not be interpreted as interchangeable indicators of a single underlying capability.

Together, these measures characterize target identification, autonomous re-engagement, and contextual continuity as distinct observable outcomes.

\section{Experimental Design}
\label{sec:experiments}

\subsection{Social Interaction Scenarios and Data Construction}
\label{subsec:social_scenarios}

We construct a benchmark consisting of 30 frozen synthetic
conversation scenarios through Microsoft Copilot, shared across
experimental conditions and evaluated models. The scenarios
cover four conversational categories: Technical/Problem-Solving
(8), Everyday/Lifestyle (8), Opinion/Discussion (7), and
Entertainment/Interest (7).

Each stimulus follows the social re-entry scenario illustrated in Figure~\ref{fig:overview}(a). An initial interaction (R1) establishes a target conversation between the agent and another participant. This is followed by five rounds of unrelated social activity, during which the target conversation remains inactive. The environment then reaches the observation state used for evaluation.

At observation, the environment contains ten candidate conversations, consisting of the original target conversation and nine distractor threads. The interruption length and number of candidate conversations are fixed across stimuli. All evaluated model calls use a temperature of zero.

The same frozen stimuli are reused across the experimental configurations, providing a common basis for comparing the three evaluation dimensions and the effects of different conditions.

\subsection{Experimental Conditions and Controls}
\label{subsec:experimental_setup}

Our experimental design considers two factors: persistent social memory and conversational resurfacing. The memory factor compares two configurations, denoted as A0 and A2. A0 omits persistent social history, although the agent may still receive task-specific information about the target conversation. In particular, the reconstruction prompt retains the target title, topic, and latest reply. A2 additionally provides recent messages from previously posted threads and the full history of the target conversation.

The resurfacing factor compares B0 and B3. Under B0, no notification is provided, and the target conversation remains in its original position among the candidates. Under B3, the agent receives a notification concerning the target conversation, which is also promoted to the first position. Thus, B3 represents a combined resurfacing intervention involving both notification and candidate ordering.

These conditions are applied differently across the three evaluation branches, as detailed in Section~\ref{subsec:evaluation_protocol}.

We include additional control conditions to examine the interpretation of the main experimental comparisons.

For Returning, the distractor-memory condition provides a memory block constructed from distractor threads under B0. This condition allows an exploratory comparison between the A0 baseline, the social history provided under A2 (including the target conversation), and a memory block containing unrelated conversational information.

The distractor-memory blocks are not strictly matched to A2. Their lengths differ from the corresponding A2 blocks by 15--22\%, exceeding the intended 5\% tolerance. They also differ in headers, thread composition, role and message distributions, and surrounding context. Accordingly, this condition is treated as an exploratory comparison rather than a controlled test isolating the contribution of target-relevant memory content.

For Reconstructing, we report an Oracle control that provides the full target-conversation context. This condition offers an additional reference for interpreting continuity scores but is not treated as a calibrated performance ceiling.

\subsection{Evaluation Protocol}
\label{subsec:evaluation_protocol}

Following the independent-branch design in Figure~\ref{fig:overview}(b), we evaluate Finding, Returning, and Reconstructing through three separate tasks. Although the branches originate from the same observation state, the information and instructions presented to the agent differ according to the evaluation objective and experimental conditions.

\paragraph{Finding.}
The Finding branch evaluates target identification under B0 and B3, with the memory configuration fixed to A0. The agent is presented with candidate conversations and explicitly instructed to select a relevant prior conversation or abstain. A successful outcome occurs when the selected conversation matches the predefined target. This comparison examines whether prioritized resurfacing improves target selection when persistent social history is absent.

\paragraph{Returning.}
The Returning branch evaluates autonomous target-directed replying under four configurations: A0B0, A0B3, A2B0, and A2B3. The agent interacts with the observation environment without an explicit instruction to return to the target conversation. Its observed action is recorded and evaluated according to the Returning criterion defined in Section~\ref{subsec:three_dimensions}. A successful outcome requires a non-empty reply directed to the target conversation. By comparing the four configurations, we examine how persistent memory and resurfacing conditions relate to autonomous target-directed replying.

\paragraph{Reconstructing.}
The Reconstructing branch evaluates contextual continuity under A0 and A2. Unlike Finding and Returning, this branch does not present a candidate feed or notification. Instead, the agent is explicitly prompted to continue the target conversation using the information available under the assigned memory configuration. The generated continuation is evaluated using an LLM-based evaluation procedure with Copilot, which assigns an ordinal contextual continuity score from 0 to 3. A score of 0 indicates a generic, unrelated, or context-unaware reply; 1 indicates topical acknowledgment without a specific prior detail; 2 indicates a continuation using at least one specific prior detail; and 3 indicates a natural extension that integrates specific history and appropriate conversational tone.

The three branches are executed independently. Consequently, an agent's selection or reply in one branch does not affect the inputs or outcomes of another branch. The resulting measurements characterize the three evaluation dimensions separately rather than representing a single end-to-end execution of social re-entry.

\subsection{Models, Implementation, and Statistical Analysis}
\label{subsec:models_implementation}

We evaluate six service/model configurations: DeepSeek-V4-Flash, GLM-4.6V-FlashX, GLM-5.1-Cloud, MiniMax-M2.7-Cloud, Qwen3.5-Cloud, and GPT-5.5-OpenModel. All six configurations are evaluated using the same frozen stimuli, experimental protocol, and outcome definitions, ensuring consistent comparisons across models. We report model identifiers as provided by their respective services. Experimental outcomes are recorded by model, stimulus, evaluation branch, and condition.

All experimental conditions were designed around the same 30 frozen stimuli, which serve as the intended units for paired comparisons. Due to incomplete or invalid API calls, the number of available outcomes varies across models and experimental conditions. The condition-specific denominators used in the analysis are reported in Table~\ref{tab:denominators}, ranging from 20 to 30 per condition. Statistical comparisons are conducted using the available outcomes, with paired analyses restricted to stimuli for which both outcomes are available.

\begin{table}[t]
\centering
\caption{Condition denominators by model. Values show the number of stimuli with available outcomes per condition. Missingness arises from incomplete or invalid API calls; Qwen3.5 experienced the highest rate of incomplete calls.}
\label{tab:denominators}
\small
\setlength{\tabcolsep}{3pt}
\begin{tabular}{@{}lrrrrrrrrrr@{}}
\toprule
Model & A0B0 & A0B3 & A2B0 & A2B3 & Distractor & Disc B0 & Disc B3 & Recon A0 & Recon A2 & Oracle \\
\midrule
DeepSeek & 30 & 28 & 27 & 29 & 30 & 30 & 29 & 28 & 28 & 30 \\
GLM-4.6V & 30 & 30 & 30 & 30 & 30 & 30 & 30 & 30 & 29 & 30 \\
GLM-5.1 & 30 & 28 & 27 & 28 & 29 & 29 & 27 & 29 & 28 & 30 \\
MiniMax & 26 & 26 & 26 & 26 & 28 & 28 & 27 & 26 & 26 & 30 \\
Qwen3.5 & 20 & 20 & 23 & 24 & 23 & 23 & 20 & 22 & 20 & 23 \\
GPT-5.5 & 30 & 30 & 30 & 30 & 30 & 30 & 30 & 30 & 30 & 28 \\
\bottomrule
\end{tabular}
\end{table}

We analyze the experimental outcomes using descriptive effect sizes and within-model statistical comparisons. Finding and Returning are evaluated as binary outcomes, while Reconstructing is evaluated using ordinal contextual continuity scores.

For Finding, we compare target identification under B0 and B3. For Returning, the main comparison examines A0B0 versus A2B0 to assess the relationship between persistent-memory configurations and autonomous target-directed replying without resurfacing. We also examine the remaining combinations of memory and resurfacing conditions. For Reconstructing, we compare the contextual continuity scores obtained under A0 and A2.

For binary outcomes, we use two-sided exact McNemar tests on paired stimuli and report Fisher exact tests as an independent-sample reference. Because each primary contrast is evaluated across six model configurations, we apply Holm--Bonferroni correction within each family of six tests. Effect sizes are reported as differences in observed success rates, expressed in percentage points.

For Reconstructing, we report descriptive continuity scores and conduct exploratory paired $t$-tests and Wilcoxon signed-rank tests using stimulus-level scores, with Holm--Bonferroni correction across the six evaluated configurations. Because contextual continuity is measured on an ordinal scale, the reported mean scores are interpreted descriptively.

Throughout the analysis, statistical comparisons are used to characterize observable differences between experimental conditions. They do not, by themselves, establish the internal mechanisms responsible for an agent's behavior. In particular, a non-significant difference is not interpreted as evidence of equivalence.

\section{Experimental Results}
\label{sec:results}

We present the experimental results across six evaluated service/model configurations, following the three dimensions of agent social re-entry: Finding, Returning, and Reconstructing. We first examine how prioritized resurfacing affects target identification, followed by the effects of persistent-memory configurations on autonomous target-directed replying. We then report the exploratory distractor-memory comparison and contextual reconstruction results, before summarizing the patterns across models.

\begin{table*}[t]
\centering
\caption{Six-system summary of the $2\times2$ behavioral design and the two probe contrasts. Binary behavioral cells show successes/available outcomes; reconstruction cells show mean continuity with available $n$. Risk Difference (RD) is expressed in percentage points.}
\label{tab:effects}
\footnotesize
\setlength{\tabcolsep}{2.5pt}
\begin{tabular}{@{}lrrrrrrrrrr@{}}
\toprule
& \multicolumn{4}{c}{Target-directed reply ($2\times2$)} & \multicolumn{3}{c}{Discovery} & \multicolumn{3}{c}{Reconstruction} \\
\cmidrule(lr){2-5}\cmidrule(lr){6-8}\cmidrule(lr){9-11}
System & A0B0 & A0B3 & A2B0 & A2B3 & B0 & B3 & RD & A0 & A2 & A0$-$A2 \\
\midrule
DeepSeek-V4 & 0/30 & 2/28 & 12/27 & 15/29 & 4/30 & 29/29 & +86.7 & 2.18 (28) & 1.79 (28) & +0.39 \\
GLM-4.6V & 1/30 & 0/30 & 16/30 & 18/30 & 16/30 & 30/30 & +46.7 & 2.00 (30) & 1.69 (29) & +0.31 \\
GLM-5.1 & 2/30 & 6/28 & 19/27 & 16/28 & 8/29 & 27/27 & +72.4 & 2.24 (29) & 1.89 (28) & +0.35 \\
MiniMax-M2.7 & 2/26 & 4/26 & 11/26 & 13/26 & 11/28 & 27/27 & +60.7 & 1.77 (26) & 1.27 (26) & +0.50 \\
Qwen3.5 & 0/20 & 1/20 & 10/23 & 10/24 & 4/23 & 20/20 & +82.6 & 2.05 (22) & 1.50 (20) & +0.55 \\
GPT-5.5 & 0/30 & 5/30 & 9/30 & 9/30 & 13/30 & 30/30 & +56.7 & 1.90 (30) & 1.83 (30) & +0.07 \\
\bottomrule
\end{tabular}
\end{table*}

\subsection{Finding: Effects of Conversational Resurfacing}
\label{subsec:results_finding}

\begin{figure}[t]
\centering
\includegraphics[width=\columnwidth]{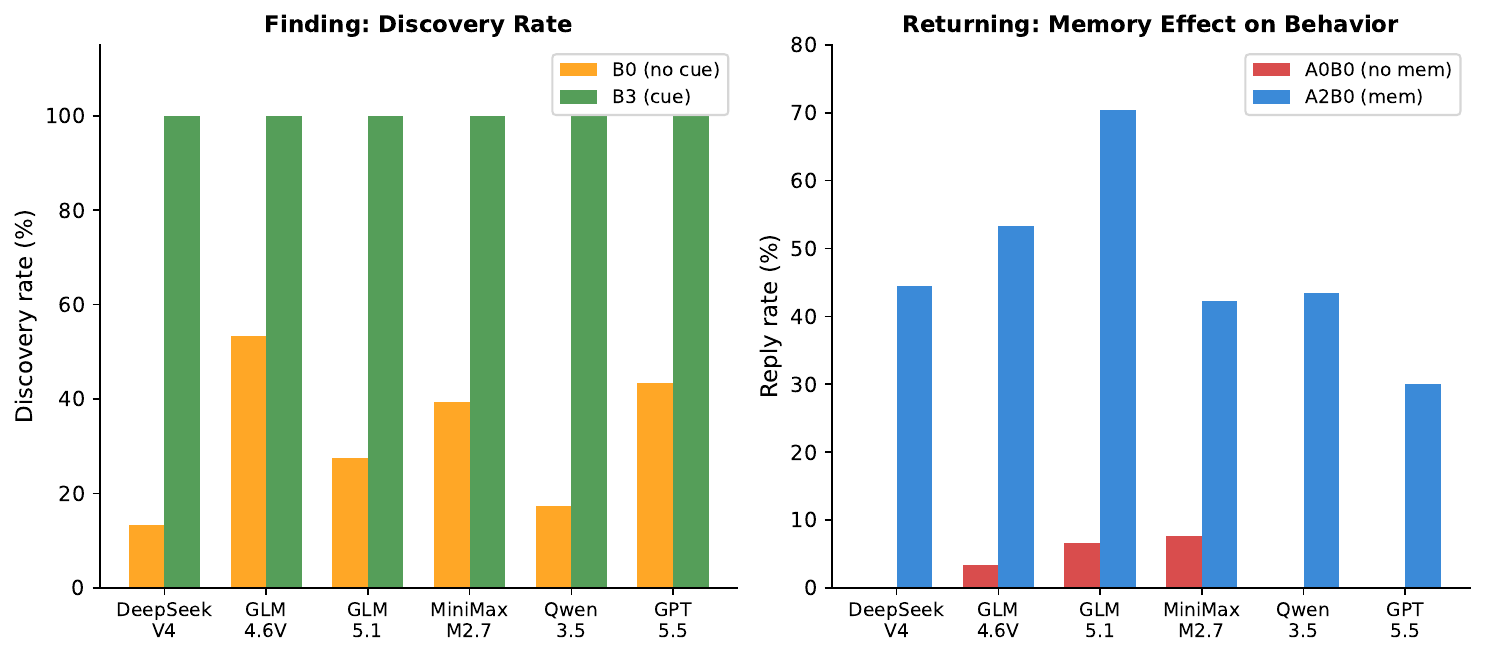}
\Description{Two side-by-side bar charts comparing six LLM configurations. The left chart shows Finding success rates under B0 and B3, with B3 reaching 100\% across all configurations. The right chart compares target-directed reply rates under A0B0 and A2B0, showing substantially higher rates with A2 memory, although these remain below the 100\% Finding success observed under B3.}
\caption{Finding versus returning across six models. Left: Discovery rates under B0 (no cue) and B3 (cue). B3 reaches 100\% in every system. Right: Target-directed reply rates under A0B0 (no memory) and A2B0 (memory). Memory is associated with large increases, but the absolute rates remain far below the 100\% discovery ceiling under B3.}
\label{fig:finding_returning}
\end{figure}

We first examine whether prioritized resurfacing improves the agent's ability to identify a previously inactive target conversation. Under B0, Finding success rates ranged from 13.3\% to 53.3\% across the six evaluated configurations. Under B3, all six configurations achieved 100\% Finding success among the available evaluations. The corresponding risk differences ranged from +46.7 to +86.7 percentage points (Table~\ref{tab:effects}; Figure~\ref{fig:finding_returning}).

The improvements were statistically significant across all six configurations. Two-sided Fisher exact tests yielded $p$-values ranging from $1.68\times10^{-5}$ to $7.84\times10^{-13}$. Exact McNemar tests on 20--30 common stimuli per configuration also showed significant differences, with all six comparisons remaining significant after Holm--Bonferroni correction (adjusted $p$-values from $3.58\times10^{-7}$ to $1.22\times10^{-4}$). Across all matched pairs, no B0 success was paired with a B3 failure.

These results show that the B3 resurfacing intervention substantially increased target identification across the evaluated configurations. As B3 combines notification with first-position promotion, the observed improvement reflects the combined intervention rather than the independent contribution of either component.

However, successful target identification did not translate into target-directed replying at a comparable rate. Under A0B3, Returning success rates ranged from 0\% to 21.4\%, compared with 100\% Finding success under B3. Specifically, the A0B3 reply rates were 7.1\%, 0\%, 21.4\%, 15.4\%, 5.0\%, and 16.7\% for DeepSeek, GLM-4.6V, GLM-5.1, MiniMax, Qwen, and GPT, respectively. This contrast illustrates the distinction between explicitly identifying a previous conversation and autonomously producing a reply to it. Since the two outcomes are measured through independent tasks, the comparison characterizes a difference between evaluation outcomes rather than a sequential failure within the same execution.

\subsection{Returning: Effects of Persistent Memory}
\label{subsec:results_returning}

We next examine how persistent-memory configurations relate to autonomous target-directed replying. Without resurfacing, the A0B0 reply rates ranged from 0.0\% to 7.7\%, whereas the A2B0 rates ranged from 30.0\% to 70.4\% across the six evaluated configurations. The corresponding risk differences were positive for every configuration, ranging from +30.0 to +63.7 percentage points (Table~\ref{tab:effects}; Figure~\ref{fig:finding_returning}).

Two-sided Fisher exact tests indicated significant differences for all six configurations ($p=.00873$ to $8.53\times10^{-7}$). Exact McNemar tests on common stimuli likewise detected significant paired differences. After Holm--Bonferroni correction, all six comparisons remained significant, with adjusted $p$-values of .0020 for DeepSeek, .0003 for GLM-4.6V, .0001 for GLM-5.1, .031 for MiniMax, .031 for Qwen, and .012 for GPT.

\begin{figure}[t]
\centering
\includegraphics[width=\columnwidth]{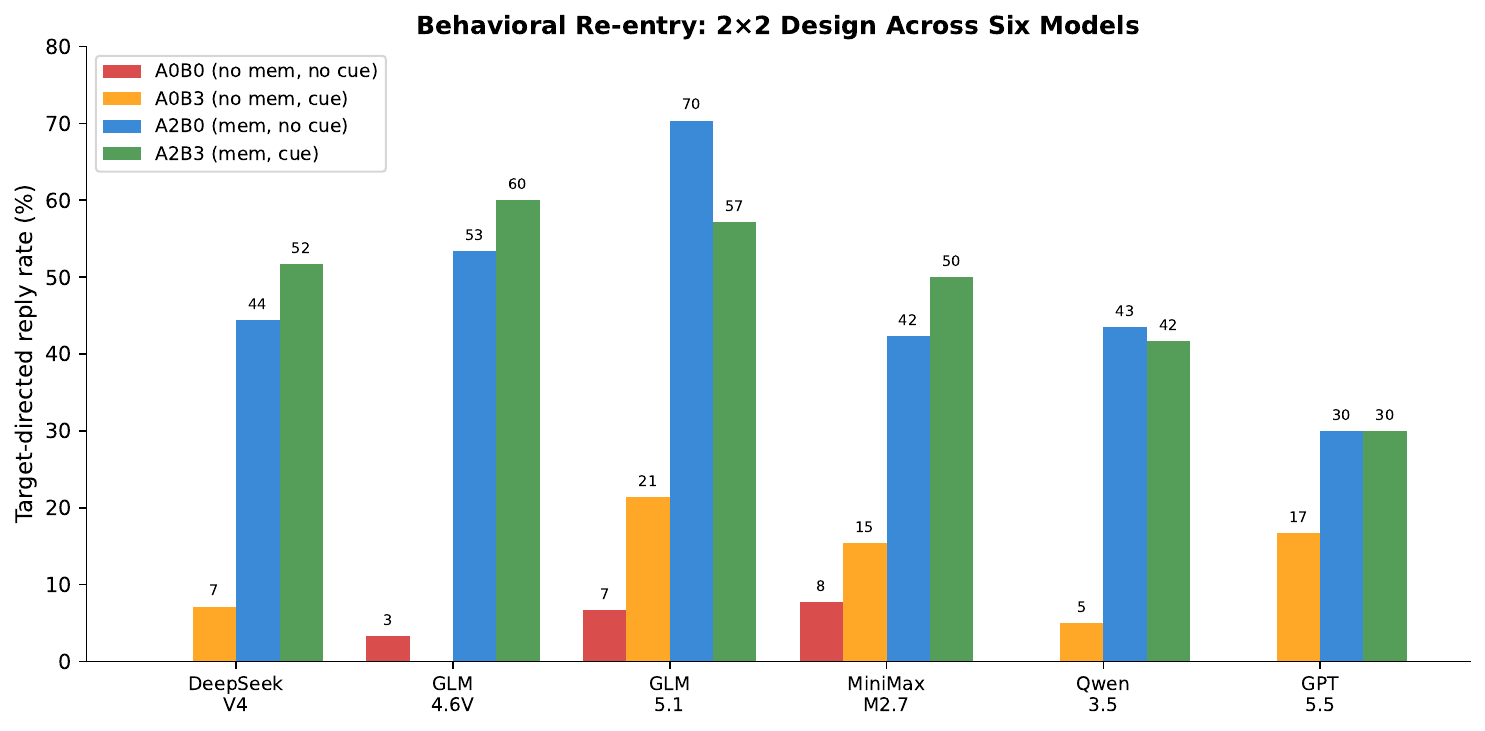}
\Description{Grouped bar chart comparing target-directed reply rates across six LLM configurations under four experimental conditions: A0B0, A0B3, A2B0, and A2B3. Each group contains four bars representing combinations of persistent memory and conversational resurfacing. The A2 memory conditions generally produce higher reply rates than A0, while the additional effect of B3 resurfacing varies across models.}
\caption{Full $2\times2$ behavioral design across six models. Each group of four bars shows A0B0, A0B3, A2B0, and A2B3 target-directed reply rates. Memory (A2) shifts rates upward; resurfacing (B3) produces a smaller and more heterogeneous additional shift.}
\label{fig:behavioral_2x2}
\end{figure}

The full $2\times2$ results further show that the effects of resurfacing on target-directed replying were smaller and more heterogeneous than the A0-to-A2 differences (Figure~\ref{fig:behavioral_2x2}). Under A0B3, reply rates remained between 0\% and 21.4\%, while both A2 configurations generally produced substantially higher rates than their A0 counterparts.

Together, these findings indicate a consistent association between the A2 memory configuration and increased autonomous target-directed replying. These differences characterize the observed association with the implemented memory configuration, rather than establishing a specific internal memory mechanism.

\subsection{Exploratory Distractor-Memory Comparison}
\label{subsec:results_distractor}

To further examine the Returning results, we compare the distractor-memory condition with the A0B0 baseline and the A2B0 configuration.

Distractor-memory reply rates ranged from 33.3\% to 60.7\% and exceeded A0B0 rates in all six evaluated configurations. Exact McNemar comparisons between distractor memory and A0B0 were significant in every configuration, with all comparisons remaining significant after Holm--Bonferroni correction (adjusted $p$-values from .0007 to .008).

\begin{figure}[t]
\centering
\includegraphics[width=\columnwidth]{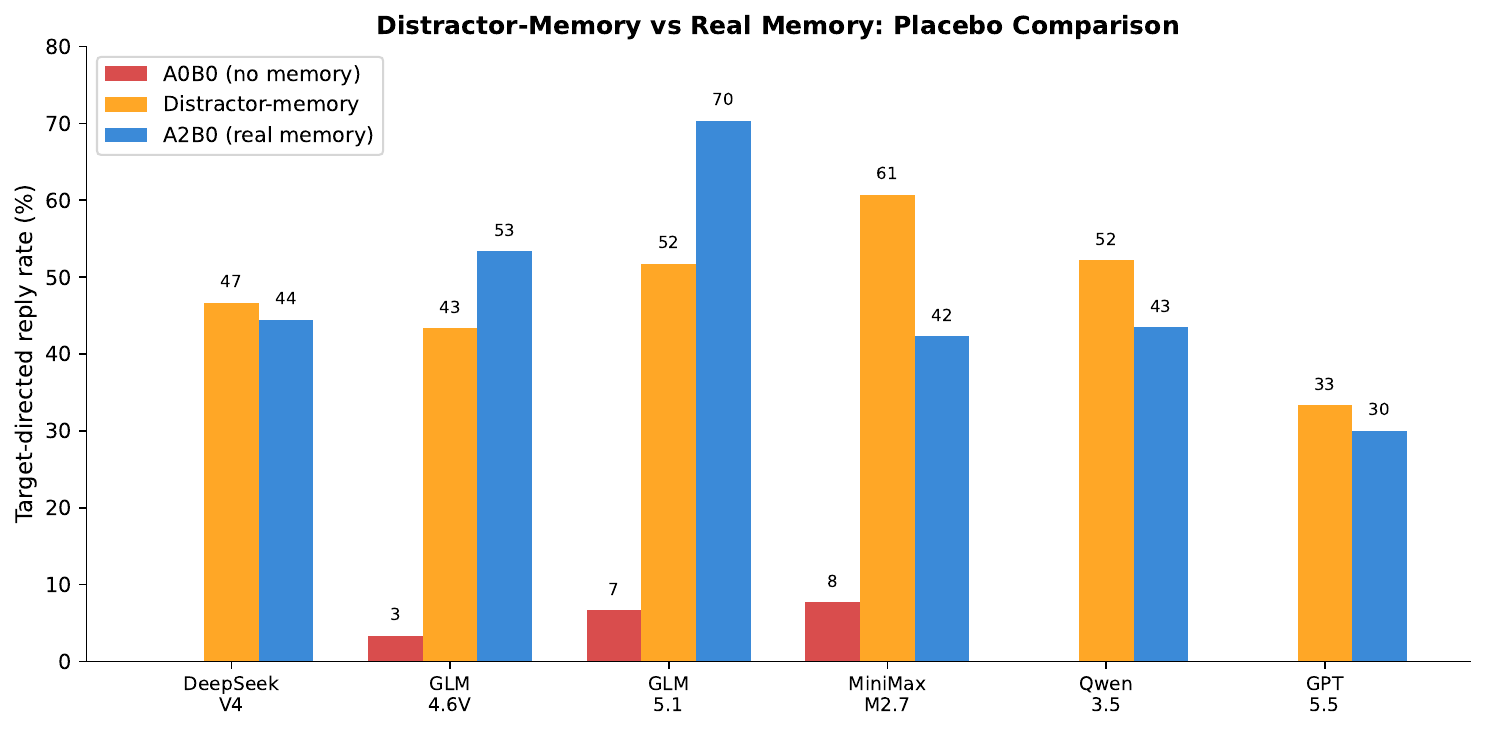}
\Description{Grouped bar chart comparing target-directed reply rates across six LLM configurations under A0B0, distractor-memory, and A2B0 conditions. Distractor-memory produces higher reply rates than A0B0 in all six configurations. Compared with A2B0, distractor-memory yields higher rates in four configurations and lower rates in two, with no consistent directional pattern.}
\caption{Distractor-memory versus real memory. Here, real memory refers to the A2B0 configuration, which includes the target conversation history, whereas distractor-memory contains unrelated conversational history. The distractor-memory condition (orange) exceeds A0B0 (red) in every system but varies around A2B0 (blue), lying above it in four systems and below it in two.}
\label{fig:placebo}
\end{figure}

In contrast, comparisons between distractor memory and A2B0 showed no consistent direction (Figure~\ref{fig:placebo}). Distractor-memory minus A2B0 risk differences ranged from $-18.6$ percentage points for GLM-5.1 to $+18.4$ percentage points for MiniMax. Distractor-memory reply rates exceeded A2B0 in four configurations and were lower in two. Exact McNemar tests did not detect a significant distractor-memory versus A2B0 difference in any configuration after Holm correction (all adjusted $p\geq .18$).

These findings show that the distractor-memory configuration was also associated with higher target-directed reply rates than A0B0. However, the absence of statistically significant differences from A2B0 does not establish equivalence. In addition, the distractor-memory and A2 configurations were not strictly matched in length and structure. Consequently, this exploratory comparison does not establish whether memory-block presence or target-relevant memory content is responsible for the observed behavioral differences.

\subsection{Reconstructing: Contextual Continuity}
\label{subsec:results_reconstructing}

We finally examine whether the A2 memory configuration improves the contextual continuity of prompted conversation continuation. Across all six evaluated configurations, the mean Reconstructing scores were lower under A2 than under A0 (Table~\ref{tab:effects}).

When calculated on common scored stimuli, the paired mean differences (A0 minus A2) were 0.46 for DeepSeek, 0.35 for GLM-4.6V, 0.26 for GLM-5.1, 0.48 for MiniMax, 0.55 for Qwen, and 0.07 for GPT. The corresponding marginal mean differences reported in Table~\ref{tab:effects} vary slightly because the numbers of available observations differ across conditions.

Exploratory paired comparisons showed heterogeneous statistical support. Only Qwen3.5 retained significance after Holm--Bonferroni correction (adjusted $p=.046$). MiniMax showed the next strongest evidence (unadjusted $p=.025$; adjusted $p=.126$), while the remaining four configurations did not exhibit statistically significant differences after correction.

\begin{figure*}[t]
\centering
\includegraphics[width=\textwidth]{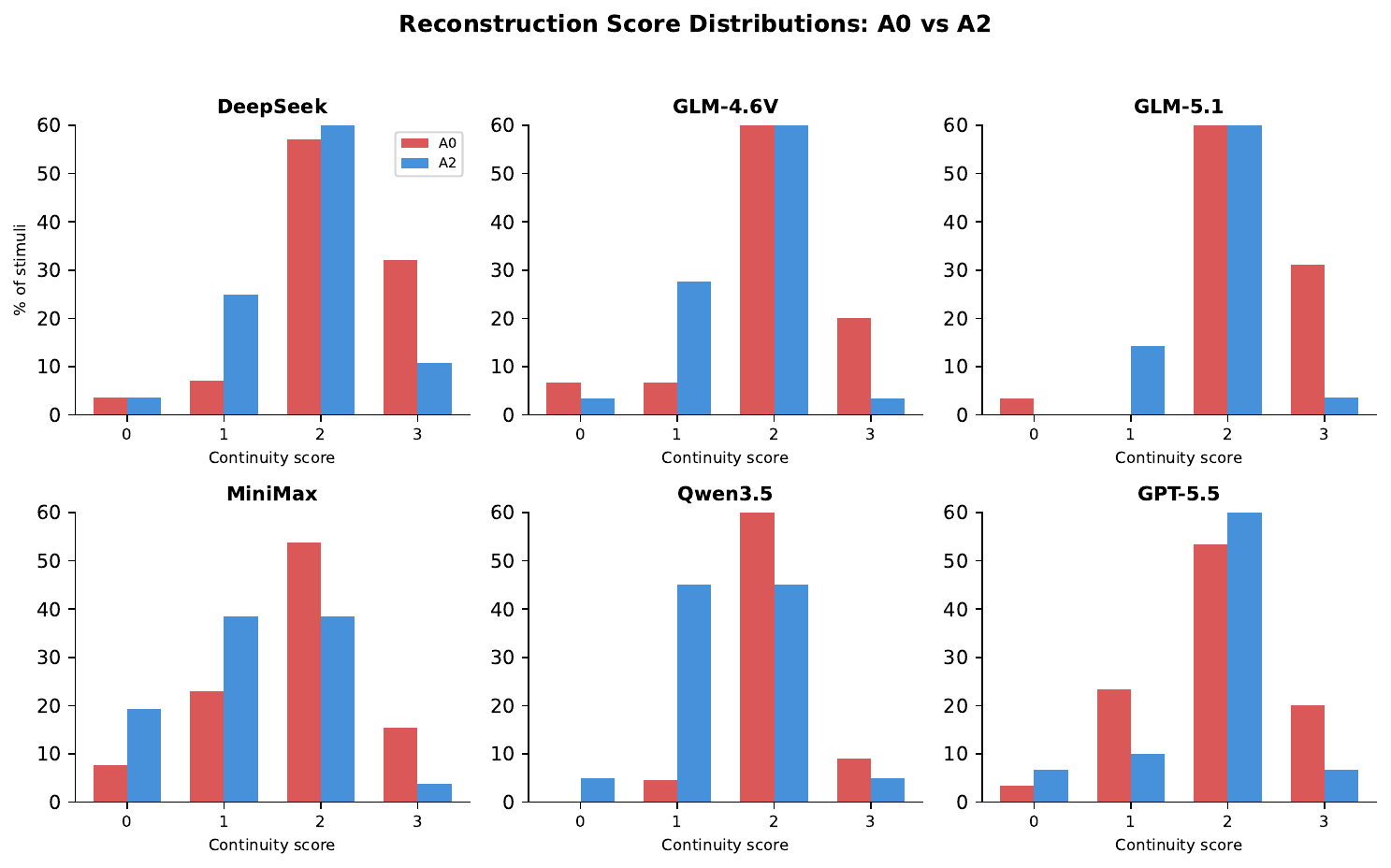}
\Description{Six panels showing the distributions of contextual continuity scores from 0 to 3 under A0 and A2 for six LLM configurations. Each panel compares the percentage of stimuli receiving each score under the two conditions. A0 generally has a higher proportion of high continuity scores than A2, with the most pronounced differences observed for MiniMax-M2.7 and Qwen3.5.}
\caption{Reconstruction score distributions (0--3 ordinal) under A0 (red) and A2 (blue) for each of the six models. Mean continuity scores are higher under A0 across all six configurations, with larger observed differences for MiniMax and Qwen3.5.}
\label{fig:reconstruction_dist}
\end{figure*}

The score distributions also show a directional shift toward higher continuity scores under A0, particularly for MiniMax and Qwen3.5 (Figure~\ref{fig:reconstruction_dist}). The percentages are calculated using the available outcomes for each model and condition, with denominators reported in Table~\ref{tab:denominators}. Nevertheless, the magnitude of the mean differences varied substantially across configurations, and the GPT-5.5 difference was only 0.07 points.

The reconstruction controls provide additional context. Oracle mean scores ranged from 1.59 to 2.03 and did not consistently exceed the corresponding A0 scores. Thus, the Oracle condition did not establish a uniform empirical performance ceiling under the implemented evaluation settings.

Overall, the results do not provide consistent evidence that A2 improves contextual continuity. Although the observed mean differences favor A0 across all six configurations, the statistical evidence does not support a uniformly negative effect of A2. Importantly, the higher target-directed reply rates associated with A2 in the Returning branch were not accompanied by corresponding improvements in the separately evaluated Reconstructing branch.

\subsection{Cross-Model Consistency}
\label{subsec:results_consistency}

\begin{figure}[t]
\centering
\includegraphics[width=\columnwidth]{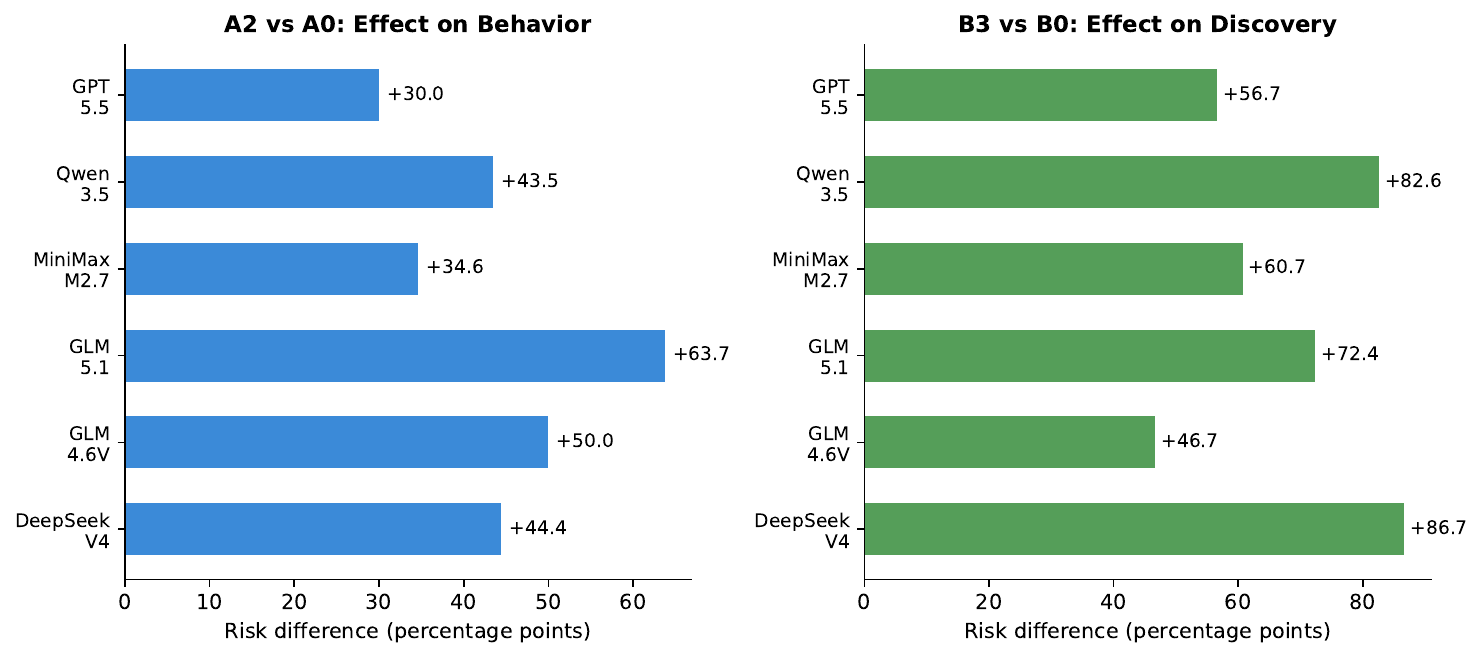}
\Description{Two side-by-side horizontal bar charts showing risk differences in percentage points across six LLM configurations. The left chart shows the increase in target-directed reply rates from A0B0 to A2B0, ranging from 30.0 to 63.7 percentage points. The right chart shows the increase in Finding success rates from B0 to B3, ranging from 46.7 to 86.7 percentage points. All risk differences are positive across the six configurations.}
\caption{Risk differences across six models. Left: A2B0 minus A0B0 for target-directed replies. Right: B3 minus B0 for discovery. All effects are positive; magnitudes vary but no model reverses either direction.}
\label{fig:effect_sizes}
\end{figure}

Across the six evaluated service/model configurations, the two primary binary outcomes exhibit consistent directional patterns. The B3 resurfacing intervention increased Finding success rates relative to B0 in every configuration, with risk differences ranging from +46.7 to +86.7 percentage points. Similarly, A2B0 produced higher target-directed reply rates than A0B0 in every configuration, with risk differences ranging from +30.0 to +63.7 percentage points (Figure~\ref{fig:effect_sizes}). Both sets of paired comparisons remained statistically significant after Holm--Bonferroni correction within their respective families of six tests.

The Reconstructing results show a different pattern. Although mean contextual continuity scores were lower under A2 in every evaluated configuration, the differences varied in magnitude, and only one configuration retained statistical significance after correction. The distractor-memory comparison was also less consistent: it exceeded A0B0 in all six configurations but varied in direction relative to A2B0.

Across the evaluated systems, resurfacing consistently improved Finding, and the A2 memory configuration was associated with higher Returning rates. However, comparable improvements in Reconstructing were not observed. These results highlight the importance of evaluating Finding, Returning, and Reconstructing separately, rather than treating success in one dimension as evidence of success in another.

\section{Discussion}
\label{sec:discussion}

\subsection{Finding, Returning, and Reconstructing as Separate Evaluation Targets}

Our results show that Finding, Returning, and Reconstructing capture
distinct aspects of agent social re-entry. Although these dimensions
are conceptually connected, improvements in one do not necessarily
correspond to improvements in another. Evaluating them separately
therefore provides a more complete picture of agent social re-entry.

The contrast between Finding and Returning is particularly
informative. Under B3, all six evaluated configurations achieved
100\% Finding success, yet target-directed replying remained
much less frequent without persistent social history. This
suggests that making a previous conversation easy to identify
does not necessarily lead an agent to return to it autonomously.
Evaluations based only on explicit conversation selection may
therefore overestimate an agent's ability to re-engage with
previous social interactions.

The contrast between Returning and Reconstructing reveals a
different limitation. The A2 memory configuration was associated
with substantially higher Returning rates across all six
configurations, but did not produce consistent improvements
in contextual continuity. Furthermore, the exploratory
distractor-memory comparison suggests that higher Returning
rates cannot be attributed solely to access to target-relevant
history. Together, these findings indicate that providing
additional conversational history may encourage autonomous
re-engagement without necessarily improving the quality of
the resulting continuation.

These findings support evaluating Finding, Returning, and
Reconstructing as separate outcomes rather than using one
as a proxy for another. Importantly, the three dimensions
are evaluated through independent experimental branches.
The observed differences do not establish a causal sequence
or imply distinct internal cognitive mechanisms.

\subsection{Implications for Persistent Social Agents}
\label{subsec:discussion_implications}

Our findings suggest that persistent social agents need to address three related but distinct challenges: identifying relevant previous conversations, deciding when to re-engage, and maintaining contextual continuity after returning. Improvements in one area should not be assumed to resolve the others.

First, conversation resurfacing should help agents identify relevant earlier interactions within an evolving social environment. Our Finding results indicate that the implemented notification-and-promotion condition substantially improved target identification. However, the contrast with Returning suggests that making an interaction salient does not necessarily produce autonomous engagement. Effective agent design therefore requires attention not only to which conversations are surfaced, but also to how an agent determines whether a reply is appropriate.

Second, persistent memory should support more than the presence of historical information in the prompt. Persistent memory should not be evaluated solely by whether it increases target-directed replies; its relevance and contribution to subsequent conversational continuity must also be considered. Although the A2 configuration was associated with increased target-directed replying, the exploratory distractor-memory comparison suggests that the relationship between memory content and observable behavior is not straightforward. Memory blocks may differ in content, length, structure, and contextual cues, each of which could influence the resulting actions. The current experiments do not identify which of these properties drives the observed differences. Future memory systems should therefore distinguish the retrieval of relevant social history from its presentation and use in action selection.

Third, contextual reconstruction warrants evaluation independently of both target discovery and replying. The absence of an observed reconstruction benefit under A2 suggests that supplying additional history does not automatically produce a more coherent continuation. Persistent social agents may require context-selection and representation strategies that prioritize information necessary to resume a particular conversational thread, rather than simply increasing the amount of history supplied.

Taken together, these findings motivate evaluation frameworks that assess whether an agent can locate an earlier interaction, autonomously act on it when appropriate, and continue it with sufficient contextual continuity. Reporting these outcomes separately can reveal limitations that would otherwise be obscured by a single social re-entry success measure.

\subsection{Limitations and Future Work}
\label{subsec:discussion_limitations}

The results suggest several directions for extending the evaluation of agent social re-entry.

One direction is to study more realistic and dynamic social environments. Our current evaluation uses controlled scenarios with fixed interruption lengths and candidate conversation sets, allowing consistent comparisons across models. Future evaluations could consider open-world settings with dynamically evolving conversations, varying interruption durations, and more diverse interaction patterns. Such settings would help examine how Finding, Returning, and Reconstructing perform under the complexity of long-term social interactions.

A second direction is to further investigate the roles of conversational resurfacing and persistent memory. Our current conditions combine multiple factors: B3 includes both notification and conversation promotion, while A2 changes the availability and presentation of conversational history. The exploratory distractor-memory comparison also involves differences in memory length and structure. Future experiments could separately examine resurfacing strategies and compare more closely matched memory configurations to better understand how memory relevance, selection, and presentation relate to social re-entry.

A third direction is to investigate how Finding, Returning, and Reconstructing relate in end-to-end agent behavior. Our independent-branch design allows the three dimensions to be evaluated separately but does not measure their progression within a single continuous interaction. Future work could extend the framework to longitudinal evaluations that trace how agents identify previous conversations, autonomously return to them, and maintain contextual continuity over time.

\section{Conclusion}
\label{sec:conclusion}

In this paper, we investigate agent social re-entry and formulate it as a multi-dimensional evaluation problem. We propose a shared-snapshot evaluation framework consisting of three complementary dimensions: Finding, Returning, and Reconstructing. The framework evaluates these dimensions independently, allowing us to examine different aspects of social re-entry without treating them as a single success outcome.

Our experiments across six LLM service/model configurations reveal that successful Finding does not necessarily imply Returning, and increased Returning does not guarantee stronger Reconstructing. While conversational resurfacing improves Finding and persistent memory is associated with higher Returning rates, these improvements are not accompanied by corresponding gains in Reconstructing. These findings highlight the distinction between the three dimensions and the importance of evaluating them separately.

More broadly, our work extends the evaluation of persistent social agents beyond memory retrieval and isolated conversational responses. By distinguishing the ability to identify previous conversations, autonomously re-engage with them, and maintain contextual continuity, our framework provides a foundation for evaluating how agents sustain social interactions over time.


\bibliographystyle{ACM-Reference-Format}
\bibliography{references}

\end{document}